%% file: bmvc_main.tex
\documentclass{bmvc2k}

\usepackage{graphicx}
\graphicspath{{figures/}}
\usepackage{booktabs}
\usepackage{tikz}
\usepackage{comment}
\usepackage{amsmath,amssymb} 
\usepackage{color}

\usepackage{floatrow}
\usepackage{caption}

\title{SearchTrack: Multiple Object Tracking with Object-Customized Search and Motion-Aware Features}

\addauthor{Zhong-Min Tsai}{vtsai01@cmlab.csie.ntu.edu.tw}{1*}
\addauthor{Yu-Ju Tsai}{r06922009@cmlab.csie.ntu.edu.tw}{1*}
\addauthor{Chien-Yao Wang}{kinyiu@iis.sinica.edu.tw}{2}
\addauthor{Hong-Yuan Liao}{liao@iis.sinica.edu.tw}{2}
\addauthor{Youn-Long Lin}{ylin@cs.nthu.edu.tw}{3}
\addauthor{Yung-Yu Chuang}{cyy@csie.ntu.edu.tw}{1}

\addinstitution{
National Taiwan University\\
Taipei, Taiwan
}
\addinstitution{
Institute of Information Science, Academia Sinica\\
Taipei, Taiwan
}

\addinstitution{
National Tsing Hua University\\
Hsinchu, Taiwan
}

\runninghead{Z. TSAI ET AL.}{SearchTrack: Multiple Object Tracking}

\begin{document}

\maketitle

\def\thefootnote{*}\footnotetext{These authors contributed equally to this work.}

\input{abstract}
\input{intro}

\input{related}
\input{searchtrack}

\input{experiments}

\input{conclusion}

\paragraph{Acknowledgments.}
This work was supported in part by MOST 110-2634-F-002-051.
We thank to National Center for High-performance Computing (NCHC) for providing computational and storage resources.

\bibliography{egbib}
\end{document}

%% file: abstract.tex
\begin{abstract}

The paper presents a new method, SearchTrack, for multiple object tracking and segmentation (MOTS). To address the association problem between detected objects, SearchTrack proposes object-customized search and motion-aware features. By maintaining a Kalman filter for each object, we encode the predicted motion into the motion-aware feature, which includes both motion and appearance cues. For each object, a customized fully convolutional search engine is created by SearchTrack by learning a set of weights for dynamic convolutions specific to the object. Experiments demonstrate that our SearchTrack method outperforms competitive methods on both MOTS and MOT tasks, particularly in terms of association accuracy. Our method achieves 71.5 HOTA (car) and 57.6 HOTA (pedestrian) on the KITTI MOTS and 53.4 HOTA on MOT17. In terms of association accuracy, our method achieves state-of-the-art performance among 2D online methods on the KITTI MOTS. Our code is available at \href{https://github.com/qa276390/SearchTrack}{https://github.com/qa276390/SearchTrack}.

\end{abstract}

%% file: intro.tex
\section{Introduction}
\label{sec:intro}

Recently, deep learning has contributed significantly to advances in various core computer vision tasks, including object detection and image segmentation. Nevertheless, there are still a number of important tasks that remain challenging.
The tracking of multiple objects is one of them, as noted by Voigtlaender~et~al\mbox{.}~\cite{Voigtlaender19CVPR_MOTS}. Multiple-object tracking (MOT) and multiple-object tracking and segmentation (MOTS) have received increased attention in recent years. MOT requires the tracking of objects using bounding boxes, while MOTS requires pixel-level accuracy. In the MOTS problem, detection, segmentation, and tracking have to be considered simultaneously~\cite{Voigtlaender19CVPR_MOTS}.

A main challenge of the MOTS task is the association of instances of the same object at different times. A number of factors make it challenging, including object deformations, changes in view, differences in illumination, occlusions, and ambiguities, among others. Object appearance and object motion are popular cues for resolving the association between detected object instances. 
The appearance cue is the most popular, and there have been many attempts to derive unique and invariant representations for objects from pixel values. 
TrackR-CNN~\cite{Voigtlaender19CVPR_MOTS} and PointTrack~\cite{xu2020Segment} are notable examples.
TrackR-CNN, based on Mask R-CNN~\cite{he2017mask}, uses region-of-interest (ROI) for cropping the feature map to generate re-identification (re-ID) feature encoding appearance cues for each instance. PointTrack converts the input frame into a 2D point cloud representation and separates the point cloud into foreground and background for learning instance embeddings.
Object motion is another popular cue for tracking, providing the object's location over time. As the tracked object in MOT(S) is highly correlated across consecutive frames, tracing its trajectory is helpful for tracking. SORT~\cite{Bewley2016sort} is a motion-based MOT tracker, which uses a Kalman filter to predict the object's motion and finds matches between objects using the Hungarian algorithm.

In this paper, we propose a new MOT(S) architecture, SearchTrack, which integrates both object appearance and motion cues to resolve the association problem. 
In order to better utilize the motion cue, the motion predicted by the Kalman filter is encoded together with the appearance feature in order to produce a motion-aware feature. In order to identify the association for a given object, we propose an object-customized search. Inspired by CondConv~\cite{yang2019condconv} and CondInst~\cite{tian2020conditional}, our model learns a set of convolution kernels for searching the given object in the current frame. The resultant convolution kernels are applied to the motion-aware feature to generate a probability map indicating the spatial likelihood of locating the given object in the current frame. To achieve good search results, dynamic convolution kernels are expected to encode the query object's characteristics, such as its appearance, relative position, and shape. Despite that similar ideas have been explored in different contexts, our main contribution is the overall framework for integrating them into the context of MOTS. Experiments show that the proposed method outperforms competitive methods on popular MOT and MOTS benchmarks, particularly in terms of association accuracy.

%% file: related.tex
\section{Related Work}

Online trackers generally focus on either motion or appearance cues, with a greater focus on the latter. Some attempts combine both cues, but they tend to do so in a simplistic manner. Methods are divided into three categories based on how they utilize cues.



\noindent \textbf{Motion Cues}.
SORT~\cite{Bewley2016sort} uses the Kalman filter~\cite{Kalman_1960} to predict object motion and estimate the future location of the tracklets. It then associates newly detected tracklets with the highest IOU overlap using the Hungarian algorithm. Since SORT only utilizes motion cues, it can achieve 260 fps inference speed and is effective in some scenarios. Given good detection results and the high frame-rate input, IOU-Tracker~\cite{1517Bochinski2017} eliminates the motion prediction algorithm and uses only the location of objects to compute overlap between the tracklets for tracking. 
In more challenging situations, such as crowded scenes, these methods may fail due to the lack of information about the appearance of the objects. 
To address the problem, Deep SORT~\cite{Wojke2017simple,Wojke2018deep} considers appearance features from a deep neural network and associates objects using the object location overlap between frames.

\noindent \textbf{Appearance Cues}.
Most trackers~\cite{Voigtlaender19CVPR_MOTS,yu2016poi,wang2020towards} extract the re-ID features from the regions proposed by a detector. FairMOT~\cite{zhang2021fairmot} shows the advantage of the center-based re-ID methods. The re-ID feature similarity is computed between tracklets and detections to assign the identities. CenterTrack~\cite{zhou2020tracking} proposes a simultaneous detection and tracking algorithm and links objects implicitly in adjacent frames using point-based object representations. SiamMOT~\cite{shuai2021siammot} formulates the Region Proposal Network and Siamese-based MOT trackers and deals with the similarity of object appearance. These methods can handle challenging cases such as identity losing and re-appearing.
CCPNet~\cite{xu2021continuous} presents a data augmentation strategy, continuous copy-paste, to deal with the limited number of instances in consecutive raw frames. However, its training uses two external datasets, and others cannot train their models using the augmentation strategy unless the pre-processed datasets become available.  

\noindent \textbf{Combination of both cues}.
In addition to appearance, FairMOT also takes into account motion cues at the time of inference. After the model has been processed, it performs the location prediction using motion cues. In a similar way to Deep SORT, FairMOT adds the similarity of two parts with a handcrafted ratio as the new similarity. 
There are two drawbacks to these methods. The first problem is that the combination ratio is fixed regardless of the size, category, and density of objects in the scene. In addition, they consider the appearance of an object and its motion as two separate sources of information.

Our approach integrates both appearance and motion cues in a more unified and learnable manner. This results in more robust association results for different scenes, object classes, and sizes, as they are adapted simultaneously. In our experiments, we found that the importance of motion differed when tracking cars and pedestrians. 
As a reasonable explanation, a car's motion is considered to be rigid body motion, whereas pedestrians' motion is considered to be non-rigid body motion. Unlike the former, which is an overall body movement, the latter is an articulated movement that is difficult to master.

%% file: searchtrack.tex
\section{SearchTrack}
\label{sec:searchtrack}

Our MOTS method, SearchTrack, is point-based and compatible with most point-based detection methods. Our current implementation is built on the CenterNet detector~\cite{zhou2019objects}, which identifies objects by their center points. For segmentation, we adapt CondInst~\cite{tian2020conditional}. On top of the point-based detector, we propose an object-customized search method to address the object association problem. For better utilizing motion cues, the Kalman filter predicts object position, which supplements the motion-aware feature. 

\begin{figure}[t]
    \centering
    \includegraphics[width=0.94\textwidth]{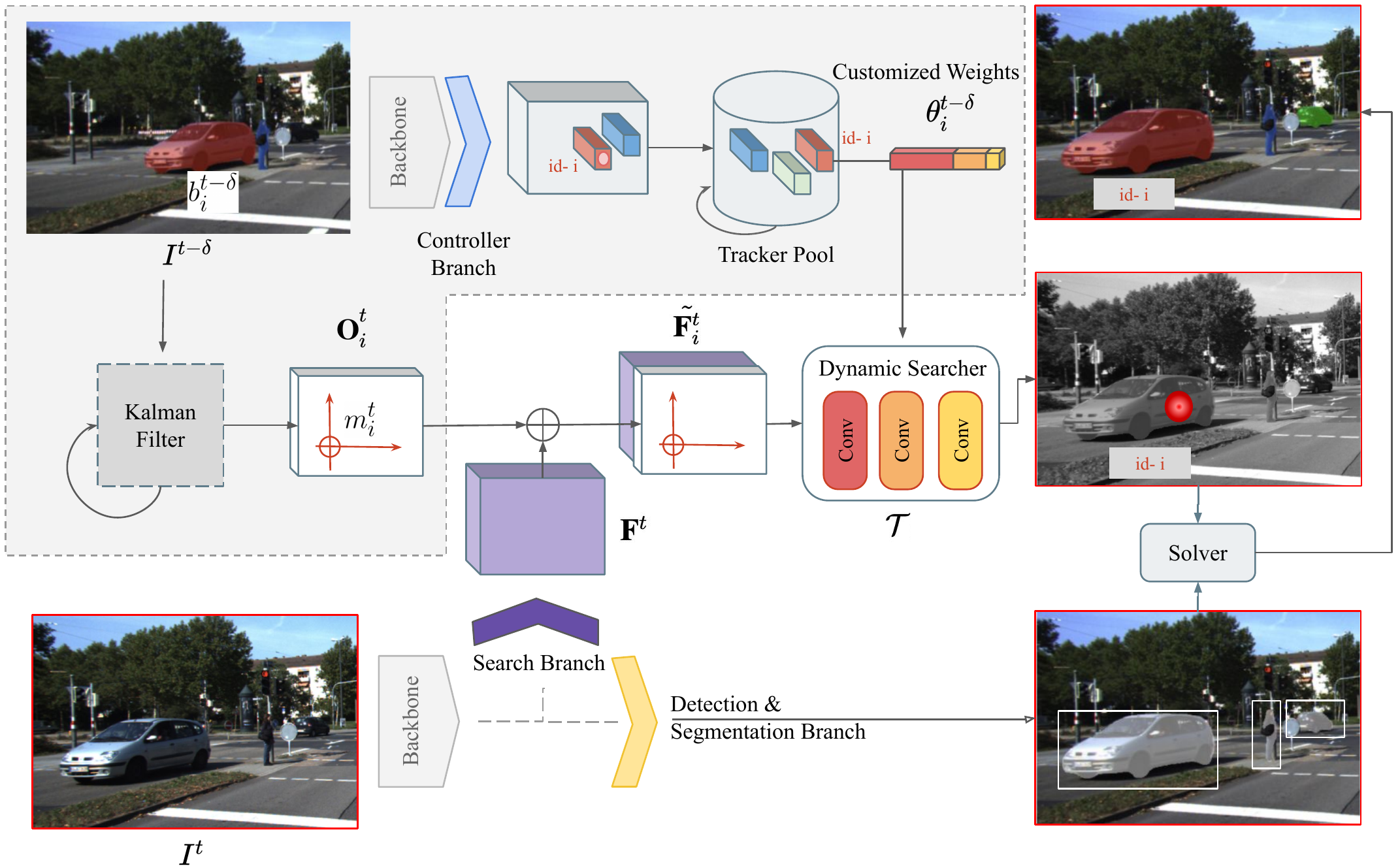}
    \caption{\textbf{Overview of SearchTrack.} 
    A shared backbone network extracts features from the current frame $I^t$ and the previous frame $I^{t-\delta}$. The extracted feature of $I^t$ is fed into the detection and segmentation branch for obtaining a set of candidate objects. It is also fed to the search branch to form a search feature map $\mathbf{F}^t$. For a query object $b_i^{t-\delta}$ in the previous frame, we obtain its customized weight $\theta_i^{t-\delta}$ and predicted location $m_i^t$ from its Kalman filter. The location $m_i^t$ is encoded in a motion map $\mathbf{O}_{i}^t$. Combining $\mathbf{O}_{i}^t$ and $\mathbf{F}^t$ gives us the motion-aware motion map $\tilde{\mathbf{F}_i^{t}}$. The customized weight $\theta_i^{t-\delta}$ realized a customized searcher $\mathcal{T}$ for the object $b_i^{t-\delta}$. By taking $\tilde{\mathbf{F}_i^{t}}$ as input, $\mathcal{T}$ outputs a response map indicating where the object could locate. Finally, a solver matches the boxes from the search branch and the candidate objects from the detection and segmentation branch for resolving association among objects. 
    }
    \label{fig:searchtrack}
\end{figure}

Fig.~\ref{fig:searchtrack} provides an overview of SearchTrack. Our method takes two images as input at a time, and these two images need not be adjacent to each other in our training process. At time $t$, an image of the current frame $I^t$ and a previous frame $I^{t-\delta}$ are given. Since the previous frame $I^{t-\delta}$ has already been processed, it contains instances with the information of tracked identities, $T^{t-\delta} = \{b^{t-\delta}_{1} ,..., b^{t-\delta}_{i}, ...\}$. Each detected object $b = (\mathbf{p}, \mathbf{m}, w, \theta,id)$ has attributes including the center location $\mathbf{p}$, the segmentation mask $\mathbf{m}$, the detection confidence $w$, the customized weights $\theta$, and the unique identity $id$. 

In the first step, both images $I^t$ and $I^{t-\delta}$ go through the same backbone network to obtain feature maps $f^t$ and $f^{t-\delta}$. The feature map $f^t$ of the current frame is then fed to the detection and segmentation branch to produce a list of newly detected instances $\tilde{T}^{t}$ (including bounding boxes and masks) without assigning identities. The core problem is to resolve the association problem between $T^{t-\delta}$ and $\tilde{T}^{t}$. For resolving this problem, the feature map $f^t$ is also fed to the search branch for extracting a compact feature map $\mathbf{F}^t$ for later search. 

For each detected object $b_i^{t-\delta}$ in the previous frame, we aim to search its position at the current frame. For better utilizing object motion, a Kalman filter is maintained separately for each detected object. For the object $b_i^{t-\delta}$, its position at the current frame is predicted by the associated Kalman filter. The position is then encoded as a motion map $\mathbf{O}_{i}^t$. By concatenating the motion map $\mathbf{O}_{i}^t$ with the feature map $\mathbf{F}^t$, we obtain the motion-aware feature map $\tilde{\mathbf{F}_i^{t}}$ for the object $b_i$ and feed it into the dynamic searcher. 

For the previous frame, its feature map $f^{t-\delta}$ is fed to the controller branch to obtain a feature map, called dynamic weight map $\theta^{t-\delta}$. Each pixel of the map contains a vector encoding a set of convolution weights customized for an object locating at that location, if there is any. For the detected object $b_i$, we obtain its customized weights $\theta_i^{t-\delta}$ and feed it to the dynamic searcher.

The dynamic searcher takes $\theta_i^{t-\delta}$ and uses it to form a set of convolution kernels as the search kernels for the object $b_i$. With these convolution kernels, the dynamic searcher $\mathcal{T}$ becomes a customized CNN for searching the object $b_i$. By passing the motion-aware feature $\tilde{\mathbf{F}^{t}}$ through $\mathcal{T}$, we obtain a response map, in which each pixel value indicates the probability that object $b_i$ appears at that pixel for the current frame. With the response map, we can find the detected instance in $\tilde{T}^{t}$, which most likely corresponds to the object $b_i$ and thus resolve the association problem.

\subsection{Detection and Segmentation Branch}
\label{sec:pointbasedetector}
Our method is point-based. The detection results of region-based detectors may cause ambiguity during training since a single rectangular region could correspond to multiple identities in a crowded scene. This shortcoming of region-based detectors was also noted by FairMOT~\cite{zhang2021fairmot}. We have therefore decided to adopt a point-based approach and build our MOTS architecture on top of a point-based detector, CenterNet~\cite{zhou2019objects} in the current implementation.
As the segmentation branch, we adapt dynamic convolution from CondInst~\cite{tian2020conditional}. 

The detection and segmentation branch takes as input a single image $I \in \mathbb{R}^{W_{I} \times H_{I} \times 3}$ and generates a detection representation set $\{( \mathbf{p_j}, \mathbf{s_j} )\}^{N-1}_j$ for each object class $c \in \{ 0, ..., C - 1\}$. The detection consists of two attributes: the center point $\mathbf{p} \in \mathbb{R}^2$ signifies the location of each object, and the size $\mathbf{s} \in \mathbb{R}^2$ gives the height and width of the object's bounding box from regression. Furthermore, the model also generates two low-resolution maps: the heat map $Y \in [0,1]^{W \times H \times C}$ and the size map $S \in [0,1]^{W \times H \times 2}$, where $W = W_{I} / R$ and $H = H_{I} / R$ with the downsampling ratio $R=4$ in our implementation. Each center of the detected object will be represented as local maximum $\mathbf{p} \in \mathbb{R}^2$ in the heat map $Y$ along with the detection confidence $ \omega= Y_{\mathbf{p}}$ and the object size $ \mathbf{s}= S_{\mathbf{p}}$, where $Y_{\mathbf{p}}$ and $S_{\mathbf{p}}$ are the values of $Y$ and $S$ at the point $p$. For segmentation, a segmentation mask is associated with each detected object.

\subsection{Motion-aware Feature Map}
\label{sec:kalmanfilter}


As mentioned, a compact search feature map $\mathbf{F}^t \in \mathbb{R}^{W \times H \times C_{search}}$ is generated by the search branch connected to the backbone branch, which takes $I^t$ as input. We empirically sets the number of channels $C_{search}$ to 16 as it achieves a good balance between computation cost and performance. 
To leverage object motion for tracking, we choose to use the Kalman filter framework~\cite{Bewley2016sort} as the motion model. We approximate the displacements between frames of each object with a linear constant velocity model, independent of the motions of other objects and the camera. For each object, a Kalman filter is maintained for predicting its locations in the following frames. By using the Kalman filter to predict the center location $m^{t}_{i}$ of the object $b^{t-\delta}_i$ at time $t$, we can produce a motion map $\mathbf{O}_{i}^t \in \mathbb{R}^{W \times H \times 2}$ that contains offset vectors from the position $m^{t}_{i}$, $\mathbf{O}_{i}^t(\mathbf{p})=\mathbf{p} - m^{t}_{i}$.
By concatenating $\mathbf{O}_{i}^t$ and $\mathbf{F}^t$, we obtain the motion-aware feature map $\tilde{\mathbf{F}_i^{t}} \in \mathbb{R}^{W \times H \times (C_{search}+2)}$ for the object $b_i$.  
The motion map provides a strong cue for predicting the object motion for the association and significantly improves the performance, particularly for non-rigid bodies. 


\subsection{Dynamic Searcher}
\label{sec:dynamicsearcher}


A core problem of multiple object tracking is to resolve association. 
SearchTrack addresses this problem by performing object-customized searches through the dynamic search engine. 
Given a detected instance $b^{t-\delta}_i $ at time $t-\delta$, its customized weights $\theta^{t-\delta}_i$ is adopted as convolution weights to search for the particular instance in frame $I^{t}$ globally. Formally,
\begin{equation} \label{eq:3-1-1}
\mathbf{R}^t_i= \mathcal{T}(\tilde{\mathbf{F}^{t}_i}; \theta^{t-\delta}_i),
\end{equation}
where $\tilde{\mathbf{F}^{t}_i}$ is the motion-aware feature map customized for the object $b_i$ and $\mathcal{T}$ is the learnable fully convolutional network (FCN) tracker with parameters $\theta^{t-\delta}_i$. These parameters are the dynamic weights generated by the controller head at the center of $b^{t-\delta}_i$. The output of this network is a response map $\mathbf{R}^t_i \in [0,1]^{W \times H}$ that gives the association possibility of $b_i$ in $I^t$. The supplementary document provides more details regarding the model architecture and the number of parameters for the dynamic weight $\theta$.

\begin{figure}[t]
    \centering
    \includegraphics[width=0.95\textwidth]{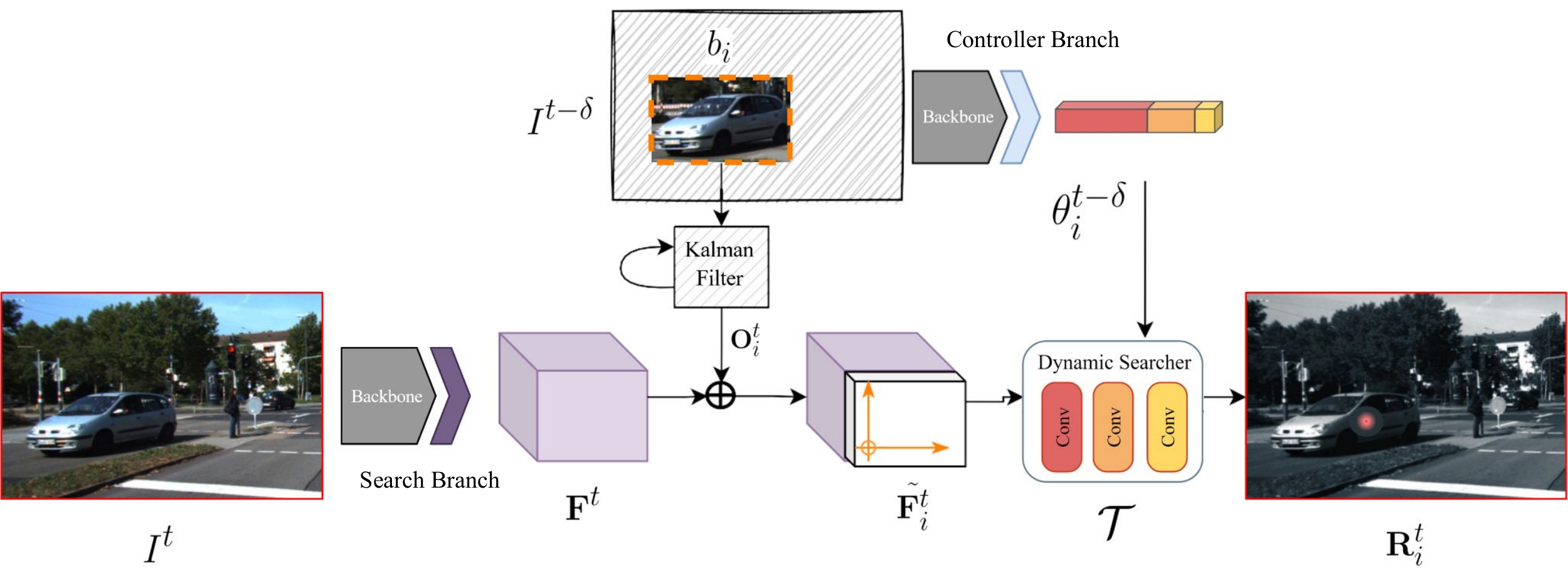}
    \caption{\textbf{The search process for finding object association}. 
    Given the input image $I^{t}$ and a detected object $b_i$ at time $t-\delta$, we obtain the feature map $\mathbf{F}^t$ from the search branch and dynamic weight $\theta^{t-\delta}_{i}$ from the controller branch. Note that both branches share the same backbone for their first parts. Next, we generate the motion map $\mathbf{O}^t_i$ according to the Kalman filter's prediction for the object.
    We then integrate $\mathbf{O}^t_i$ and $\mathbf{F}^t$ to have the motion-aware feature map $\tilde{\mathbf{F}^{t}_i}$ for the object. 
    The dynamic weight $\theta^{t-\delta}_{i}$ is fed into the dynamic searcher $\mathcal{T}$ to realize a customized searcher for the given object. By taking $\tilde{\mathbf{F}^{t}_i}$ as input, $\mathcal{T}$ outputs an association response map $\mathbf{R}^t_i$ for the object in the current frame. 
    The peak of the map indicates the center of the object $b_i$ in $I^t$.
    }
    \label{fig:track}
\end{figure}

By finding the peak in the response map, we determine the location of $b^{t}_i$ and take the peak value as the association confidence $\upsilon^t_i$. Note that the instance $b^{t}_i$ is not completed until the bounding box size and segmentation branch is applied.
In the case that the instance $b^{t-\delta}_i$ is visible in $I^t$, our tracker $\mathcal{T}$ should give a high confidence score at the location where $b^t_i$ is located. Fig.~\ref{fig:track} illustrates the association process in the overall model depicted in Fig.~\ref{fig:searchtrack}.


\begin{figure}[t]
    \centering
    \includegraphics[width=0.95\textwidth]{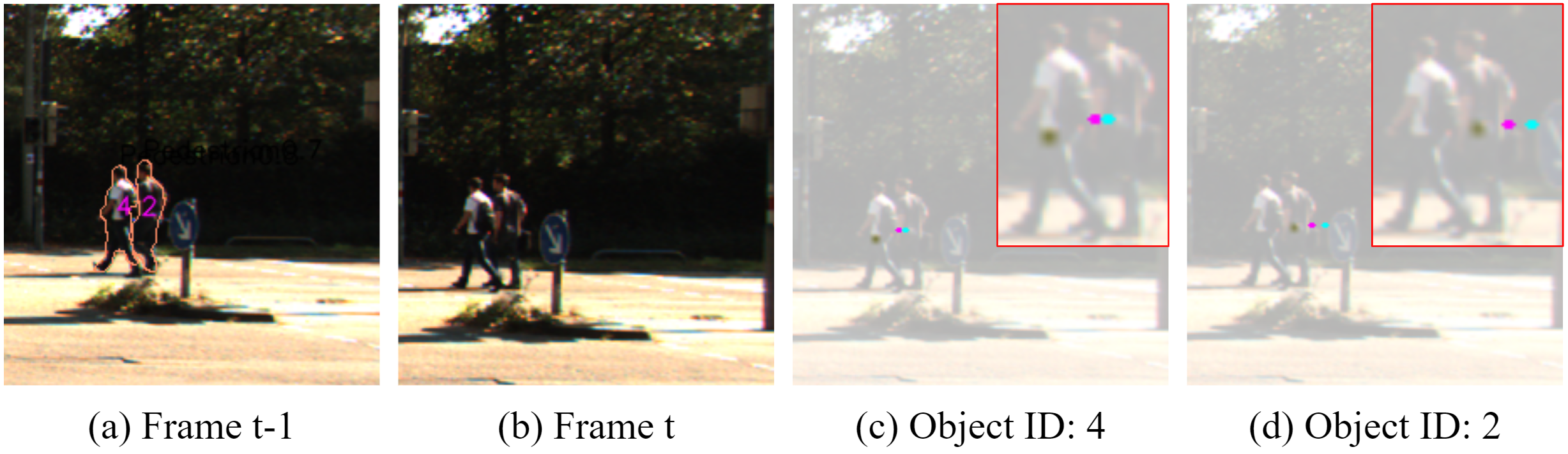}
    \caption{\textbf{The response maps from the dynamic searcher.} 
    (a) The tracking and segmentation results from the previous frame at time $t-1$.
    (b) The current frame at time $t$.
    (c) The response map for the object with ID 4 at time $t$.
    (d) The response map for the object with ID 2 at time $t$.
    There are two detected objects with ID 2 and 4 at frame $t-1$ as shown in (a). 
    The response map of a previously detected object indicates the likelihood that the object appears at each pixel at time $t$.
    Taking the object with ID 4 as an example, the green intensity in the response map (c) shows the corresponding likelihood of the object. 
    The peak will be taken as the detected position of the object. 
    For (c) and (d), the cyan point indicates the object center in the previous frame at time $t-1$. The magenta point denotes the predicted object center provided by the Kalman Filter at time $t$. For clarity, the relevant region is enlarged and shown on the top-right corner.
    }
    \label{fig:responsemap}
\end{figure}

Fig.~\ref{fig:responsemap} gives an example of the response map. Given the objects in the previous frame and the current frame, the searcher generates a response map corresponding to each object. The green intensity value of each pixel indicates the probability that the object appears at that pixel. The cyan points indicate the previous object centers, and the magenta points represent the object center predicted by the Kalman Filter. 

The procedure in Eq. (\ref{eq:3-1-1}) is applied multiple times, once for each detected object $b^{t-\delta}_i$. Nevertheless, each frame only needs to pass through the backbone network one time, resulting in a significant reduction in computation time. 
Without using many parameters, the dynamic searcher shows that a compact FCN with dynamically-generated filters can outperform ROI-based trackers such as TrackR-CNN.


\subsection{Loss Function}
\label{sec:loss}
The overall loss function of SearchTrack can be formulated as:
\begin{equation} \label{eq:3-3-1}
L_{total} = L_{heatmap} + L_{mask} + L_{search},
\end{equation}
where $L_{mask}$ is the loss of the CondInst segmentation branch; $L_{heatmap}$ is the focal loss~\cite{law2018cornernet,lin2017focal} defined in the CenterNet,
\begin{equation} \label{eq:3-0}
L_{heatmap} = \frac{1}{N} \sum_{xyc} 
\begin{cases} 
(1-{Y}_{xyc})^\alpha log({Y}_{xyc}) &\mbox{if } Y_{xyc}^* = 1
\\ 
(1 - Y_{xyc}^*)^\beta ({Y}_{xyc})^\alpha log(1-{Y}_{xyc}) &\mbox{otherwise,}
\end{cases}
\end{equation}
where $Y \in [0,1]^{W \times H \times C}$ is the output heatmap and $Y^* \in [0,1]^{W \times H \times C}$ is the ground-truth heatmap corresponding to the objects. $N$ is the number of objects, and $\alpha=2$ and $\beta=4$ are hyperparameters of the focal loss. 

We define the loss $L_{search}$ as
\begin{equation} \label{eq:3-3-2}
L_{search} = \sum_{i} L_{focal}( \mathcal{T}(\tilde{\mathbf{F}^{t}_i}; \theta^{t-\delta}_i), {R^*}^{t}_i)
\end{equation}
and $L_{focal}$ is a reduced version of Eq. (\ref{eq:3-0}) defined as:
\begin{equation} \label{eq:3-3-3}
L_{focal} (R, R^*) = \sum_{xy} 
\begin{cases} 
(1-{R}_{xy})^\alpha log({R}_{xy}) &\mbox{if } R_{xy}^* = 1
\\ 
(1 - R_{xy}^*)^\beta ({R}_{xy})^\alpha log(1-{R}_{xy}) &\mbox{otherwise,}
\end{cases}
\end{equation}
where $R^* \in [0,1]^{W \times H}$ is the ground truth response map corresponding to the objects. The ground truth map is formed by rendering a Gaussian at the true center position. $\alpha = 2$ and $\beta = 4$ are hyperparmeters of the focal loss. 


%% file: experiments.tex
\begin{table}[t]
    \centering
    \resizebox{1.0\linewidth}{!}{%
        \begin{tabular}{c  c  c  c  c  c  c  c  c  c  c c c c c} 
            \toprule[1.5pt]
            & & & & \multicolumn{5}{c}{Pedestrian} & & \multicolumn{5}{c}{Car} \\
             & & time & & {HOTA$\uparrow$}  & {DetA$\uparrow$} & {AssA$\uparrow$} & {LocA$\uparrow$} & {sMOTSA$\uparrow$} & & {HOTA$\uparrow$} & {DetA$\uparrow$}  & {AssA$\uparrow$} & {LocA$\uparrow$} & {sMOTSA$\uparrow$} \\ 
            \cline{0-0} \cline{3-3} \cline{5-9} \cline{11-15}
            TrackRCNN~\cite{Voigtlaender19CVPR_MOTS}
            & & 0.5
            & & 41.9  & 53.8  & 33.8   & 78.0  & 47.3 
            & & 56.6  & 69.9  & 46.5   & 86.6  & 67.0 
            \\
            GMPHD\_SAF~\cite{song2020online}
            & & \textcolor{blue}{0.08}
            & & 49.3  & \textcolor{blue}{65.5}  & 38.3   & \textcolor{blue}{83.8}  & \textcolor{blue}{62.9} 
            & & 55.1  & 77.0  & 39.8   & \textcolor{blue}{88.7}  & 75.4 
            \\
            PointTrack~\cite{xu2020Segment}
            & & \textcolor{red}{0.05}
            & & 54.4  & 62.3  & 48.1   & 83.3  & 61.5 
            & & 62.0  & \textcolor{red}{79.4}  & 48.8   & 88.5  & \textcolor{red}{78.5} 
            \\
            ReMOTS*~\cite{yang2021remots}
            & & 3
            & & \textcolor{red}{58.8}  & \textcolor{red}{68.0}  & \textcolor{blue}{52.4}   & \textcolor{red}{84.2}  & \textcolor{red}{66.0} 
            & & \textcolor{red}{71.6}  & \textcolor{blue}{78.3}  & \textcolor{blue}{66.0}   & \textcolor{red}{89.3}  & \textcolor{blue}{75.9} 
            \\
            \cline{0-0} \cline{3-3} \cline{5-9} \cline{11-15}
            Ours    
            & & 0.19
            & & \textcolor{blue}{57.6}  & 63.7  &  \textcolor{red}{53.1}   &  80.9 & 60.6  
            & & \textcolor{blue}{71.5}  & 76.8  &  \textcolor{red}{67.1}   &  88.0 & 74.9 
            \\
            \toprule[1.5pt]
        \end{tabular}
    }
    \caption{\textbf{Comparison with leading 2D methods on the KITTI MOTS leaderboard.}\\ ``*'' denotes \textbf{offline} methods. The ``time'' column reports inference time (in seconds) for a single frame, which was reported by the authors. We compare the published leading \textbf{online} 2D methods and one \textbf{offline} 2D method on the leaderboard. Our method achieves the highest HOTA score among all \textbf{online} 2D methods. Also, our association accuracy (AssA) outperforms all the compared methods. We highlight \textcolor{red}{the best} and \textcolor{blue}{the second best} in each column.}
    \label{tab:ktmots}
\end{table}
\begin{figure}[h]
    \centering
    \includegraphics[width=0.98\textwidth]{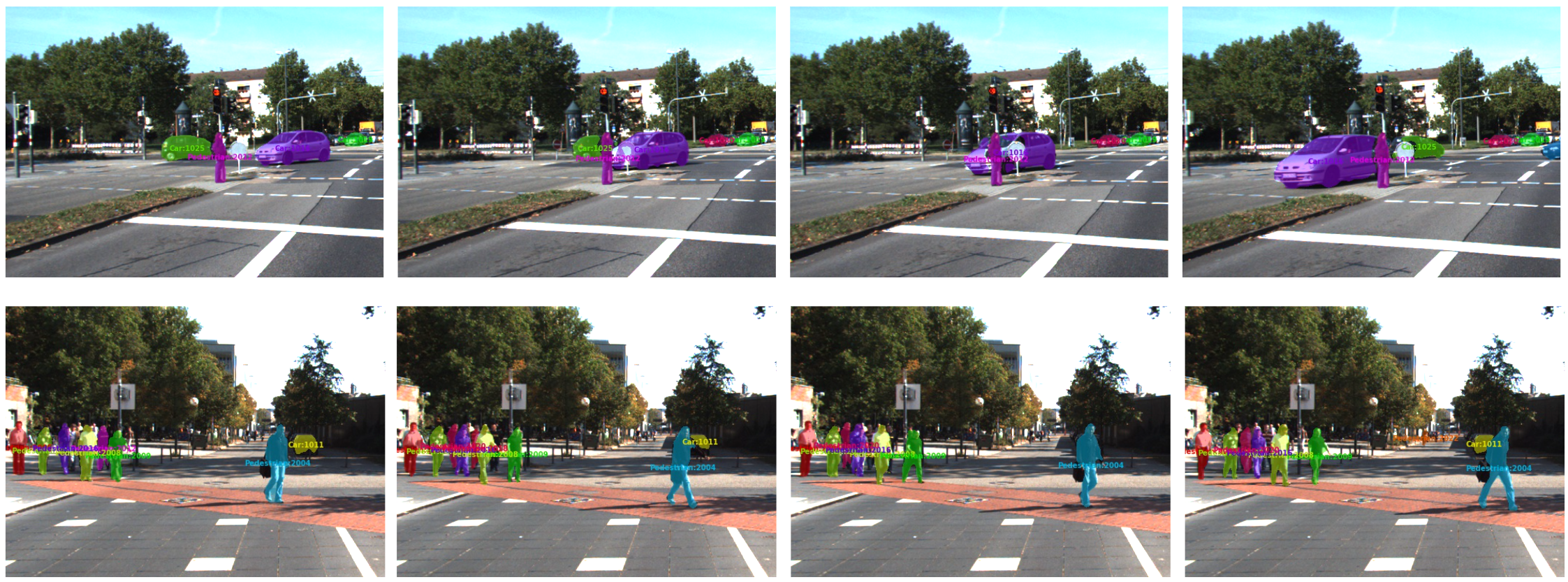}
    \caption{\textbf{Results of SearchTrack on KITTI MOTS}. Tracking ids are coded in colors.}
    \label{fig:MOTSvisual}
\end{figure}

\section{Experiments and Results}

We evaluate the proposed tracker on both the MOT and MOTS tasks using a popular MOTS benchmark, KITTI MOTS~\cite{Voigtlaender19CVPR_MOTS}, and a popular MOT benchmark, MOT17~\cite{milan2016mot16}. 
The primary evaluation metric is HOTA~\cite{luiten2020IJCV,luiten2020trackeval}, which is the \textbf{primary metric} in KITTI MOTS and MOT17 for comparisons and ranking since 2021. Additionally, we consider other metrics when appropriate, including sMOTSA, MOTA, IDF1, DetA, AssA, and LocA. The supplementary document provides more details regarding the datasets and metrics. 



\subsection{Results}


\noindent \textbf{KITTI MOTS}.
Table~\ref{tab:ktmots} compares SearchTrack with some competitive methods on the KITTI MOTS test set. In order to make a fair comparison, we only select the leading \textbf{published} 2D methods since some of the leading methods on the leaderboard are 3D and utilize additional information. 
The ``time'' column of Table~\ref{tab:ktmots} reports the average inference time (in seconds) for each method on a single frame. As far as inference time is concerned, our method is also competitive. Fig.~\ref{fig:MOTSvisual} shows some visual results.

Our SearchTrack outperforms all the state-of-the-art online 2D methods in both pedestrian and car categories. 
Our method outperforms the state-of-the-art PointTrack method~\cite{xu2020Segment} significantly: for the pedestrian class, our method provides 3.2\% gain on HOTA and 5\% gain on AssA; and for the car class, our method provides 9.5\% gain on HOTA and 18.3\% gain on AssA. In addition to the online methods, SearchTrack achieves comparable performance to the offline 2D method ReMOTS~\cite{yang2021remots} (0.7 \% gain for pedestrian and 1.1\% gain for cars on AssA) with a significant speed advantage (0.19s vs. 3s per frame). The results show that our method has significantly improved object association for tracking with a reasonable inference speed.



\begin{figure}[t]
    \centering
    \includegraphics[width=0.98\textwidth]{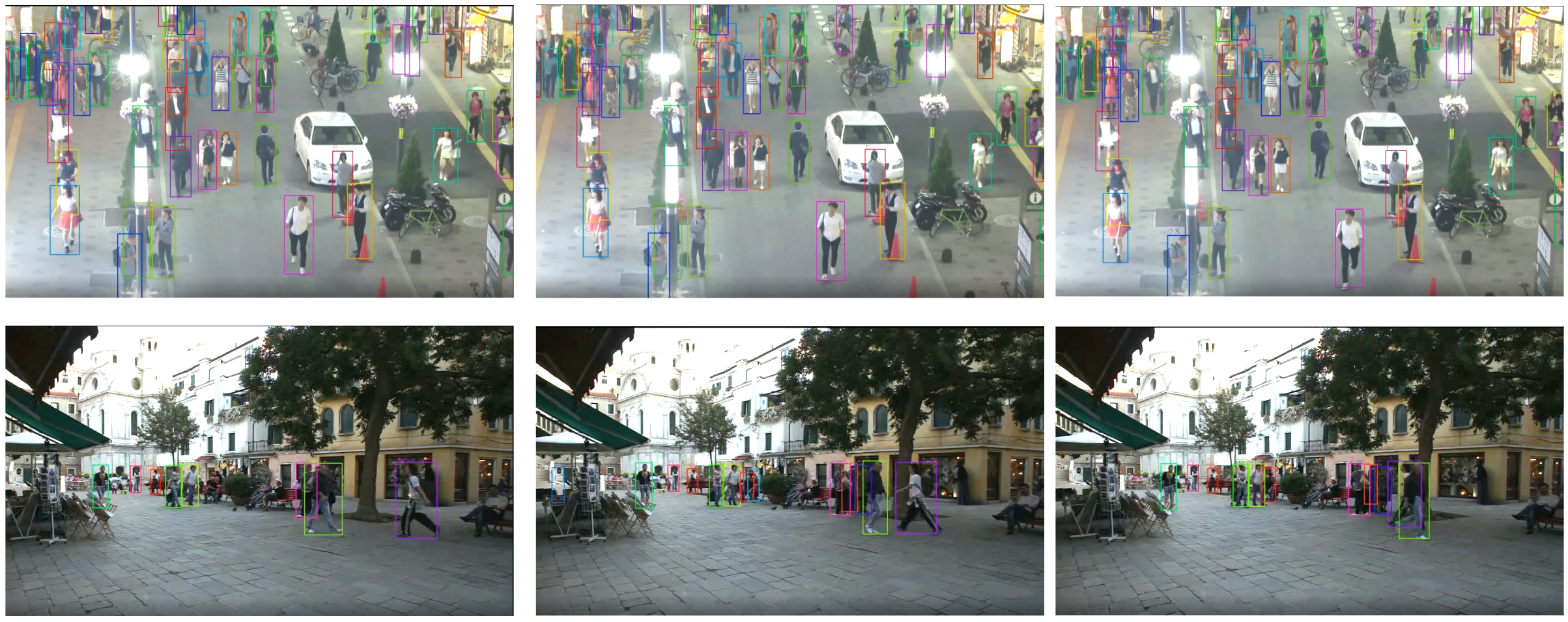}
    \caption{\textbf{Results of SearchTrack on MOT17}. Tracking ids are coded in colors.}
    \label{fig:MOTvisual}
\end{figure}

\noindent \textbf{MOT17}.
Table~\ref{tab:mot17} compares our method with the methods on the MOT17 leaderboard with the public detection setting. The MOT17 public detection setting provides the detection results and mainly tests the tracker's ability to associate objects for tracking. We use the public detection configuration with the same setting as CenterTrack and pretrain on the CrowdHuman dataset~\cite{shao2018crowdhuman}. For this comparison, although SearchTrack is 0.8 behind PermaTrack~\cite{tokmakov2021learning} on HOTA, it provides the superior association accuracy performance as measured by AssA. Our tracker is competitive with state-of-the-art methods on the MOT task. Fig.~\ref{fig:MOTvisual} shows some tracking results on the MOT17 dataset.

\begin{table}[t]
    \centering
    \resizebox{0.995\linewidth}{!}{%
        \begin{tabular}{c  c  c  c  c  c  c  c  c  c  c c c c c} 
            \toprule[1.5pt]
            {Method} & HOTA$\uparrow$  & DetA$\uparrow$ & AssA$\uparrow$ & MOTA$\uparrow$ & IDF1$\uparrow$ & MT$\uparrow$ & ML$\downarrow$ & FP$\downarrow$ & FN$\downarrow$ & IDSW$\downarrow$ \\ 
            \toprule[1pt]
            Tracktor++~\cite{2019tracktor}  & 44.8 & 44.9 & 45.1 & 56.3 & 55.1 & 21.1\% & 35.3 \% & \textcolor{red}{8866} & 235449 & 1987\\
            Visual-Spatial~\cite{bastani2021selfsupervised}  & 46.4 & 45.3 & 47.9 & 56.8 & 58.3 & 22.8\% & 37.4 \% & 11567 & 230645 & \textcolor{red}{1320}\\
            CenterTrack~\cite{zhou2020tracking}  & 48.2 & 49.0 & 47.8 & 61.5 & 59.6 & 26.4\% & 31.9 \% & 14076 & 200672 & 2583\\
            TMOH~\cite{Stadler_2021_CVPR}        & 50.4 & 49.6 & 50.9 & 62.1 & 62.8 & 26.9\% & 31.4 \% & \textcolor{blue}{10951} & 201195 & \textcolor{blue}{1897}\\
            SiamMOT~\cite{shuai2021siammot}      & - & -  & - & 65.9  & 63.3  & 34.6\% & 23.9\% & 18098 & 170955 & 3040 \\
            PermaTrack~\cite{tokmakov2021learning}   & \textcolor{red}{54.2}  & \textcolor{red}{58.0}  & \textcolor{blue}{51.2}  & \textcolor{red}{73.1}  & \textcolor{red}{67.2} & \textcolor{red}{42.3\%} & \textcolor{red}{19.1\%} & 24557 & \textcolor{red}{123508} & 3571 \\
            \toprule[1pt]
            Ours         & \textcolor{blue}{53.4}  & \textcolor{blue}{55.6} & \textcolor{red}{51.6}  & \textcolor{blue}{68.0} & \textcolor{blue}{65.7} & \textcolor{blue}{39.1\%} & \textcolor{blue}{21.1\%} & 25651 & \textcolor{blue}{150786} & 4254\\
            \toprule[1.5pt]
        \end{tabular}}
    \caption{\textbf{Results on the MOT17 test set with public detection.} We compare leading published \textbf{online} methods on the leaderboard. The results show that SearchTrack achieves the best association accuracy (AssA) among all the compared trackers. We highlight \textcolor{red}{the best} and \textcolor{blue}{the second best} in each column.}
    \label{tab:mot17}
\end{table}

\subsection{Ablation Studies}

We have conducted ablation studies to justify our main design choices in the proposed architecture, including the motion-aware feature and the segmentation branch. The experiments are mainly performed on KITTI MOTS. The supplementary gives more ablation studies.


\noindent \textbf{Motion-aware feature}. 
Table~\ref{tab:ab-motion} compares the feature maps with and without encoding motion information on the KITTI MOTS validation set. The feature map without motion information only takes appearance cues into account. Its association performance is weak, especially for the pedestrian class. It shows that using only appearance cues is often insufficient for a non-rigid body such as a pedestrian. The motion-aware feature map significantly boosts the association accuracy for the pedestrian class.

\newcommand{\bftab}{\fontseries{b}\selectfont}
\begin{table}[t]
    \centering
    \resizebox{0.9\linewidth}{!}{%
        \begin{tabular}{c c  c  c  c  c  c  c  c  c c} 
            \hline
            & & \multicolumn{3}{c}{Pedestrian} & & \multicolumn{3}{c}{Car} \\
            & & {HOTA$\uparrow$}  & {DetA$\uparrow$} & {AssA$\uparrow$} & & {HOTA$\uparrow$} & {DetA$\uparrow$} & {AssA$\uparrow$}\\ 
            \cline{0-0} \cline{3-5} \cline{7-9}
            feature without motion & & 
            43.9 \% & \bftab 56.4 \% & 34.4 \% & & 77.1 \% & 79.3 \% & 75.5 \%\\
            motion-aware feature & & 
            \bftab 59.4 \% &  56.0 \% & \bftab 63.5 \% & & \bftab 78.7 \% & \bftab 79.4 \% & \bftab 78.5 \%\\
            \hline
            
        \end{tabular}}
    \caption{\textbf{Ablation study on the motion-aware feature using the KITTI MOTS validation set.} We evaluate the importance of encoding motion into the feature map for search. The results show the motion information is important, especially for the pedestrian class. With the motion-aware feature, the association accuracy (AssA) and HOTA are significantly improved, especially for non-rigid pedestrians. }
    \label{tab:ab-motion}
\end{table}



            
            

\noindent \textbf{The segmentation branch}.
To investigate the synergy between tracking and segmentation, we experiment with training with and without the segmentation branch.
For analysis, we evaluate in bounding box level on KITTI MOTS and train from scratch, avoiding the impacts from the pre-trained model. 
Table~\ref{tab:ab-seg} shows that tracking performance is significantly improved for most evaluation metrics when augmenting the segmentation branch into architecture. The results validate that tracking can benefit from the dense pixelwise annotations and better separation of foreground and background. 
Note that we do not enable the segmentation branch in our architecture when performing the MOT task.

\begin{table}[t]
    \centering
    \resizebox{0.96\linewidth}{!}{%
        \begin{tabular}{c  c  c  c  c  c  c  c  c  c  c } 
            \hline
            & & \multicolumn{4}{c}{Pedestrian} & & \multicolumn{4}{c}{Car} \\
            
             & & {HOTA$\uparrow$}  & {DetA$\uparrow$} & {AssA$\uparrow$} & {LocA$\uparrow$} & & {HOTA$\uparrow$} & {DetA$\uparrow$}  & {AssA$\uparrow$} & {LocA$\uparrow$}\\ 
            \cline{0-0} \cline{3-6} \cline{8-11}
            w/o segmentation & & 55.9 \% & 55.5 \% & 56.4 \%  & 82.2 \% &  & 73.9 \% & \bftab 73.7 \% & 74.3 \%  & 87.3 \% \\
            w/ segmentation & & \bftab 58.2 \% & \bftab 56.8 \% & \bftab 59.8 \%  & \bftab 83.8 \% & & \bftab 74.5 \% & 73.3 \% & \bftab 76.0 \%  & \bftab 87.6 \% \\
            \hline

        \end{tabular}}
    \caption{\textbf{Impact of the segmentation branch on tracking using the KITTI MOTS validation set.} By jointly training with the segmentation task, the tracking performance can be improved. Furthermore, because the backbone is shared for these tasks, the knowledge of foreground and background also benefits the tracking task.}
    \label{tab:ab-seg}
\end{table}


%% file: conclusion.tex
\section{Conclusion}

This paper proposes an online point-based tracker named SearchTrack that simultaneously considers object appearance and motion cues to address the MOTS problem. To resolve the association problem, SearchTrack employs an object-customized search network. Also, by maintaining a Kalman filter for each object, we encode the predicted location into the motion-aware feature map as the input to the customized searcher. SearchTrack outperforms the state-of-the-art online 2D methods on the KITTI MOTS benchmark. Additionally, our method outperforms previous methods in terms of association accuracy on the MOT benchmark. As far as we know, SearchTrack is the first one-stage point-based MOTS framework. Moreover, it is efficient, straightforward, and more accurate than the state-of-the-art methods.

%% file: bmvc_main.bbl
\begin{thebibliography}{29}
\providecommand{\natexlab}[1]{#1}
\providecommand{\url}[1]{\texttt{#1}}
\expandafter\ifx\csname urlstyle\endcsname\relax
  \providecommand{\doi}[1]{doi: #1}\else
  \providecommand{\doi}{doi: \begingroup \urlstyle{rm}\Url}\fi

\bibitem[Bastani et~al.(2021)Bastani, He, and
  Madden]{bastani2021selfsupervised}
Favyen Bastani, Songtao He, and Samuel Madden.
\newblock Self-supervised multi-object tracking with cross-input consistency.
\newblock \emph{Advances in Neural Information Processing Systems}, 2021.

\bibitem[Bergmann et~al.(2019)Bergmann, Meinhardt, and
  Leal-Taixe]{2019tracktor}
Philipp Bergmann, Tim Meinhardt, and Laura Leal-Taixe.
\newblock Tracking without bells and whistles.
\newblock \emph{2019 IEEE/CVF International Conference on Computer Vision
  (ICCV)}, 2019.

\bibitem[Bewley et~al.(2016)Bewley, Ge, Ott, Ramos, and
  Upcroft]{Bewley2016sort}
Alex Bewley, Zongyuan Ge, Lionel Ott, Fabio Ramos, and Ben Upcroft.
\newblock Simple online and realtime tracking.
\newblock In \emph{2016 IEEE International Conference on Image Processing
  (ICIP)}, 2016.

\bibitem[Bochinski et~al.(2017)Bochinski, Eiselein, and
  Sikora]{1517Bochinski2017}
Erik Bochinski, Volker Eiselein, and Thomas Sikora.
\newblock High-speed tracking-by-detection without using image information.
\newblock In \emph{International Workshop on Traffic and Street Surveillance
  for Safety and Security at IEEE AVSS 2017}, 2017.

\bibitem[He et~al.(2017)He, Gkioxari, Doll{\'a}r, and Girshick]{he2017mask}
Kaiming He, Georgia Gkioxari, Piotr Doll{\'a}r, and Ross Girshick.
\newblock Mask r-cnn.
\newblock In \emph{Proceedings of the IEEE international conference on computer
  vision}, 2017.

\bibitem[Jonathon~Luiten(2020)]{luiten2020trackeval}
Arne~Hoffhues Jonathon~Luiten.
\newblock Trackeval.
\newblock \url{https://github.com/JonathonLuiten/TrackEval}, 2020.

\bibitem[Kalman(1960)]{Kalman_1960}
Rudolph~Emil Kalman.
\newblock A new approach to linear filtering and prediction problems.
\newblock \emph{Journal of Basic Engineering}, 1960.

\bibitem[Law and Deng(2018)]{law2018cornernet}
Hei Law and Jia Deng.
\newblock Cornernet: Detecting objects as paired keypoints.
\newblock In \emph{Proceedings of the European conference on computer vision
  (ECCV)}, 2018.

\bibitem[Lin et~al.(2017)Lin, Goyal, Girshick, He, and
  Doll{\'a}r]{lin2017focal}
Tsung-Yi Lin, Priya Goyal, Ross Girshick, Kaiming He, and Piotr Doll{\'a}r.
\newblock Focal loss for dense object detection.
\newblock In \emph{Proceedings of the IEEE international conference on computer
  vision}, 2017.

\bibitem[Luiten et~al.(2020)Luiten, Osep, Dendorfer, Torr, Geiger,
  Leal-Taix{\'e}, and Leibe]{luiten2020IJCV}
Jonathon Luiten, Aljosa Osep, Patrick Dendorfer, Philip Torr, Andreas Geiger,
  Laura Leal-Taix{\'e}, and Bastian Leibe.
\newblock Hota: A higher order metric for evaluating multi-object tracking.
\newblock \emph{International Journal of Computer Vision}, 2020.

\bibitem[Milan et~al.(2016)Milan, Leal-Taix{\'e}, Reid, Roth, and
  Schindler]{milan2016mot16}
Anton Milan, Laura Leal-Taix{\'e}, Ian Reid, Stefan Roth, and Konrad Schindler.
\newblock Mot16: A benchmark for multi-object tracking.
\newblock \emph{arXiv preprint arXiv:1603.00831}, 2016.

\bibitem[Shao et~al.(2018)Shao, Zhao, Li, Xiao, Yu, Zhang, and
  Sun]{shao2018crowdhuman}
Shuai Shao, Zijian Zhao, Boxun Li, Tete Xiao, Gang Yu, Xiangyu Zhang, and Jian
  Sun.
\newblock Crowdhuman: A benchmark for detecting human in a crowd.
\newblock \emph{arXiv preprint arXiv:1805.00123}, 2018.

\bibitem[Shuai et~al.(2021)Shuai, Berneshawi, Li, Modolo, and
  Tighe]{shuai2021siammot}
Bing Shuai, Andrew Berneshawi, Xinyu Li, Davide Modolo, and Joseph Tighe.
\newblock Siammot: Siamese multi-object tracking.
\newblock In \emph{CVPR}, 2021.

\bibitem[Song and Jeon(2020)]{song2020online}
Young-min Song and Moongu Jeon.
\newblock Online multi-object tracking and segmentation with gmphd filter and
  simple affinity fusion.
\newblock \emph{arXiv preprint arXiv:2009.00100}, 2020.

\bibitem[Stadler and Beyerer(2021)]{Stadler_2021_CVPR}
Daniel Stadler and Jurgen Beyerer.
\newblock Improving multiple pedestrian tracking by track management and
  occlusion handling.
\newblock In \emph{Proceedings of the IEEE/CVF Conference on Computer Vision
  and Pattern Recognition (CVPR)}, 2021.

\bibitem[Tian et~al.(2020)Tian, Shen, and Chen]{tian2020conditional}
Zhi Tian, Chunhua Shen, and Hao Chen.
\newblock Conditional convolutions for instance segmentation.
\newblock In \emph{European Conference on Computer Vision}, 2020.

\bibitem[Tokmakov et~al.(2021)Tokmakov, Li, Burgard, and
  Gaidon]{tokmakov2021learning}
Pavel Tokmakov, Jie Li, Wolfram Burgard, and Adrien Gaidon.
\newblock Learning to track with object permanence.
\newblock In \emph{ICCV}, 2021.

\bibitem[Voigtlaender et~al.(2019)Voigtlaender, Krause, Osep, Luiten, Sekar,
  Geiger, and Leibe]{Voigtlaender19CVPR_MOTS}
Paul Voigtlaender, Michael Krause, Aljosa Osep, Jonathon Luiten, Berin
  Balachandar~Gnana Sekar, Andreas Geiger, and Bastian Leibe.
\newblock {MOTS}: Multi-object tracking and segmentation.
\newblock In \emph{CVPR}, 2019.

\bibitem[Wang et~al.(2020)Wang, Zheng, Liu, Li, and Wang]{wang2020towards}
Zhongdao Wang, Liang Zheng, Yixuan Liu, Yali Li, and Shengjin Wang.
\newblock Towards real-time multi-object tracking.
\newblock In \emph{European Conference on Computer Vision}, 2020.

\bibitem[Wojke and Bewley(2018)]{Wojke2018deep}
Nicolai Wojke and Alex Bewley.
\newblock Deep cosine metric learning for person re-identification.
\newblock In \emph{2018 IEEE Winter Conference on Applications of Computer
  Vision (WACV)}, 2018.

\bibitem[Wojke et~al.(2017)Wojke, Bewley, and Paulus]{Wojke2017simple}
Nicolai Wojke, Alex Bewley, and Dietrich Paulus.
\newblock Simple online and realtime tracking with a deep association metric.
\newblock In \emph{2017 IEEE International Conference on Image Processing
  (ICIP)}, 2017.

\bibitem[Xu et~al.(2019)Xu, Meng, Yang, and Huang]{xu2021continuous}
Zhenbo Xu, Ajin Meng, Wei Yang, and Liusheng Huang.
\newblock Continuous copy-paste for one-stage multi-object tracking and
  segmentation.
\newblock In \emph{Proceedings of the IEEE/CVF International Conference on
  Computer Vision}, 2019.

\bibitem[Xu et~al.(2020)Xu, Zhang, Tan, Yang, Huang, Wen, Ding, and
  Huang]{xu2020Segment}
Zhenbo Xu, Wei Zhang, Xiao Tan, Wei Yang, Huan Huang, Shilei Wen, Errui Ding,
  and Liusheng Huang.
\newblock Segment as points for efficient online multi-object tracking and
  segmentation.
\newblock In \emph{Proceedings of the European Conference on Computer Vision
  (ECCV)}, 2020.

\bibitem[Yang et~al.(2019)Yang, Bender, Le, and Ngiam]{yang2019condconv}
Brandon Yang, Gabriel Bender, Quoc~V Le, and Jiquan Ngiam.
\newblock Condconv: Conditionally parameterized convolutions for efficient
  inference.
\newblock \emph{Advances in Neural Information Processing Systems}, 2019.

\bibitem[Yang et~al.(2020)Yang, Chang, Dang, Zheng, Sakti, Nakamura, and
  Wu]{yang2021remots}
Fan Yang, Xin Chang, Chenyu Dang, Ziqiang Zheng, Sakriani Sakti, Satoshi
  Nakamura, and Yang Wu.
\newblock Remots: Self-supervised refining multi-object tracking and
  segmentation.
\newblock \emph{arXiv preprint arXiv:2007.03200}, 2020.

\bibitem[Yu et~al.(2016)Yu, Li, Li, Liu, Shi, and Yan]{yu2016poi}
Fengwei Yu, Wenbo Li, Quanquan Li, Yu~Liu, Xiaohua Shi, and Junjie Yan.
\newblock Poi: Multiple object tracking with high performance detection and
  appearance feature.
\newblock In \emph{European Conference on Computer Vision}, 2016.

\bibitem[Zhang et~al.(2021)Zhang, Wang, Wang, Zeng, and Liu]{zhang2021fairmot}
Yifu Zhang, Chunyu Wang, Xinggang Wang, Wenjun Zeng, and Wenyu Liu.
\newblock Fairmot: On the fairness of detection and re-identification in
  multiple object tracking.
\newblock \emph{International Journal of Computer Vision}, 2021.

\bibitem[Zhou et~al.(2019)Zhou, Wang, and Kr{\"a}henb{\"u}hl]{zhou2019objects}
Xingyi Zhou, Dequan Wang, and Philipp Kr{\"a}henb{\"u}hl.
\newblock Objects as points.
\newblock In \emph{arXiv preprint arXiv:1904.07850}, 2019.

\bibitem[Zhou et~al.(2020)Zhou, Koltun, and
  Kr{\"a}henb{\"u}hl]{zhou2020tracking}
Xingyi Zhou, Vladlen Koltun, and Philipp Kr{\"a}henb{\"u}hl.
\newblock Tracking objects as points.
\newblock \emph{ECCV}, 2020.

\end{thebibliography}
